\pdfoutput=1

\documentclass[11pt]{article}

\usepackage[]{emnlp2021}

\usepackage{times}
\usepackage{latexsym}

\usepackage[T1]{fontenc}

\usepackage[utf8]{inputenc}

\usepackage{microtype}
\usepackage{latexsym}
\usepackage{amssymb}
\usepackage{amsmath}
\usepackage{amsfonts}
\usepackage{amsbsy}
\usepackage{tikz}
\usepackage{xcolor}
\usepackage{paralist}
\usepackage{float}
\usepackage{multicol}
\usepackage{multirow}
\usetikzlibrary{arrows, positioning, shapes.geometric, tikzmark, calc}

\title{Knowledge Base Completion Meets Transfer Learning}

\author{Vid Kocijan\textsuperscript{1}, Thomas Lukasiewicz\textsuperscript{1,2} \\
  \textsuperscript{1}University of Oxford \\
  \textsuperscript{2}Alan Turing Institute, London \\
  \texttt{firstname.lastname@cs.ox.ac.uk} \\}
  
\begin{document}
\maketitle
\begin{abstract}
The aim of \emph{knowledge base completion} is to predict unseen facts from existing facts in knowledge bases.
In this work, we introduce the first approach for transfer of knowledge from one collection of facts to another without the need for entity or relation matching.
The method works for both \emph{canonicalized} knowledge bases and \emph{uncanonicalized} or \emph{open knowledge bases}, i.e., knowledge bases where more than one copy of a real-world entity or relation may exist.
Such knowledge bases are a natural output of automated information extraction tools that extract structured data from unstructured text. 
%
Our main contribution is a method that can make use of a large-scale pre-training on facts, collected from unstructured text, to improve predictions on structured data from a specific domain.
The introduced method is the most impactful on small datasets such as ReVerb20K, where we obtained $6\%$ absolute increase of mean reciprocal rank and $65\%$ relative decrease of mean rank over the previously best method, despite not relying on large pre-trained models like \textsc{Bert}.
\end{abstract}
\section{Introduction}

A knowledge base (KB) is a collection of facts, stored and presented in a structured way that allows a simple use of the collected knowledge for applications.
In this paper, a \emph{knowledge base} is a finite set of triples $\langle h,r,t\rangle$, where \emph{h} and \emph{t} are \emph{head} and \emph{tail} entities, while \emph{r} is a binary relation between them.
Manually constructing a knowledge base is tedious and requires a large amount of labor.
To speed up the process of construction, facts can be extracted from unstructured text automatically, using, e.g., open information extraction (OIE) tools, such as ReVerb~\cite{reverb}  or more recent neural approaches~\cite{supervisedoie,oie_comparison}. 
Alternatively, missing facts can be inferred from existing ones using \emph{knowledge base completion (KBC)} algorithms, such as ConvE~\cite{conve}, TuckER~\cite{tucker}, or 5$^\star$E~\cite{5stare}. 

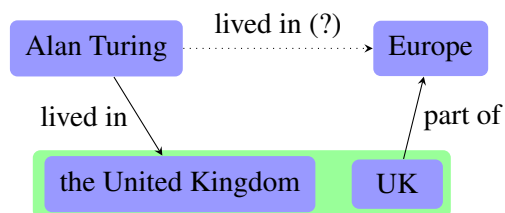
\begin{figure}
    \centering
    \begin{tikzpicture}[
        xscale = 2.2,
        yscale = 0.9,
        Block/.style = {
            fill = green!40,
            text = black,
            inner sep = 2mm,
            rounded corners = 1mm
        },
        Vect/.style = {
            fill = blue!40,
            text = black,
            inner sep = 2mm,
            rounded corners = 1mm
        },
        Arrow/.style = {
            ->,
            >=stealth
        }
    ]
    \node[Block, minimum width = 55mm, minimum height=9mm ] at (0.87,0) (inBlock){};
    \node[Vect, minimum width = 12mm] at (0.5,0) (UK_full){the United Kingdom};
    \node[Vect, minimum width = 12mm] at (1.8,0) (UK){UK};
    \node[Vect, minimum width = 12mm] at (0,2) (Alan){Alan Turing};
    \node[Vect, minimum width = 12mm] at (2,2) (Europe){Europe};
    \draw[Arrow] (Alan) -- node [left,midway] {lived in} (UK_full);
    \draw[Arrow] (UK) -- node [right,midway] {part of} (Europe);
    \draw[Arrow,dotted] (Alan) -- node [above,midway] {lived in (?)} (Europe);
    \end{tikzpicture}
    \caption{An example of a small knowledge base with the fact whether Alan Turing lived in Europe missing. If the knowledge base is canonicalized, ``the United Kingdom'' and ``UK'' are known to be the same entity. If the knowledge base is uncanonicalized or open, this information may not be given.}
    \label{task_diagram}
\end{figure}

It is desirable to use both OIE and knowledge base completion approaches to automatically construct KBs.
However, automatic extractions from text yield uncanonicalized entities and relations.
An entity such as ``the United Kingdom'' may also appear as ``UK'', and a relation such as ``located at'' may also appear as ``can be found in''.
If we fail to connect these occurrences and treat them as distinct entities and relations, the performance of KBC algorithms drops significantly~\cite{care}.
If our target data are canonicalized, collecting additional uncanonicalized data from unstructured text is not guaranteed to improve the performance of said models.
An illustration of a knowledge base can be found on Figure~\ref{task_diagram}.

Open knowledge base completion (OKBC) aims to mitigate this problem and to predict unseen facts even when real-world entities and relations may appear under several different names.
Existing work in the area overcomes this either by learning a mapping from entity and relation names to their embeddings~\cite{olpbench} or by using external tools for knowledge base canonicalization~\cite{cesi}, and using the  obtained predictions to enhance the embeddings of the entities~\cite{care,okgit}.

In this work, we follow the first of these two approaches. 
We pre-train RNN-based encoders that encode entities and relations from their textual representations to their embeddings jointly with a KBC model on a large OKBC benchmark.
We use this pre-trained KBC model and encoders to initialize the final model that is later fine-tuned on a smaller dataset.
More specifically, KBC parameters that are shared among all inputs are used as an initialization of the same parameters of the fine-tuned model.
When initializing the input-specific embeddings, we introduce and compare two approaches:
Either the pre-trained entity and relation encoders are also used and trained during the fine-tuning, or they are used in the beginning to compute the initial values of all entity and relation embeddings, and then dropped.

We evaluate our approach with three different KBC models and on five datasets, showing consistent improvements on most of them.
We show that pre-training turns out to be particularly helpful on small datasets with scarce data by achieving SOTA performance on the ReVerb20K and ReVerb45K OKBC datasets~\cite{care,cesi} and consistent results on the larger KBC datasets FB15K237 and WN18RR~\cite{fb15k237,conve}.
Our results imply that even larger improvements can be obtained by pre-training on a larger corpus.
The code used for the experiments is available at \url{https://github.com/vid-koci/KBCtransferlearning}.

We highlight that \citet{care} and \citet{olpbench} use the term ``open knowledge graph embeddings'' for OKBC.
We use the term ``open knowledge base completion'' to also include KBs with non-binary relations, which cannot be viewed as graphs, and to consider methods that may not produce embeddings.

Our main contributions are briefly as follows:
\begin{compactitem}
    \item We introduce a novel approach for the transfer of knowledge between KBC models that works on both open and regular knowledge bases without the need for entity or relation matching.
    \item We show that pre-training on a large OKBC corpus improves the performance of these models on both KBC and OKBC datasets.
    \item We obtain improvements over state-of-the-art approaches on $3$ of $5$ observed datasets, with the difference being particularly significant on the smallest datasets (e.g., $0.058$ absolute increase of MRR and $65\%$ MR decrease on the ReVerb20K dataset).
\end{compactitem}

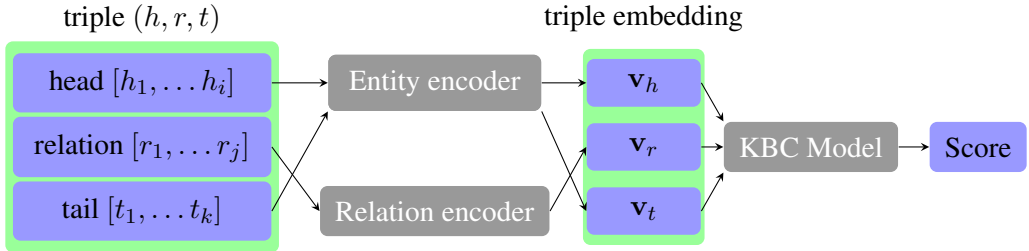
\begin{figure*}[!ht]
    \centering
    \begin{tikzpicture}[
        xscale = 2.2,
        yscale = 0.85,
        Block/.style = {
            fill = green!40,
            text = black,
            inner sep = 2mm,
            rounded corners = 1mm
        },
        Vect/.style = {
            fill = blue!40,
            text = black,
            inner sep = 2mm,
            rounded corners = 1mm
        },
        Model/.style = {
            fill = black!40,
            text = white,
            inner sep = 2mm,
            rounded corners = 1mm
        },
        Arrow/.style = {
            ->,
            >=stealth
        }
    ]
    \node[Block, minimum width = 36mm, minimum height=28mm ] at (0,1) (inBlock){};
    \node[Vect, minimum width = 34mm] at (0,0) (t){tail $[t_1, \ldots t_k]$};
    \node[Vect, minimum width = 34mm] at (0,1) (r){relation $[r_1, \ldots r_j]$};
    \node[Vect, minimum width = 34mm] at (0,2) (h){head $[h_1, \ldots h_i]$};
    \node[] at (0,3) (inputText){triple $(h,r,t)$};
    \node[Model, minimum width = 28mm] at (1.75,2) (enc1){Entity encoder};
    \node[Model, minimum width = 28mm] at (1.75,0) (enc2){Relation encoder};
    \draw[Arrow] (h) -- (enc1);
    \draw[Arrow] (t.east) -- (enc1.south west);
    \draw[Arrow] (r.east) -- (enc2.west);
    \node[Model, minimum width = 28mm] at (1.75,2) (enc1){Entity encoder};
    \node[] at (3,3) (inputText){triple embedding};
    \node[Block, minimum width = 16mm, minimum height=26mm ] at (3,1) (embBlock){};
    \node[Vect, minimum width = 15mm] at (3,0) (vt){$\mathbf{v}_t$};
    \node[Vect, minimum width = 15mm] at (3,1) (vr){$\mathbf{v}_r$};
    \node[Vect, minimum width = 15mm] at (3,2) (vh){$\mathbf{v}_h$};
    \draw[Arrow] (enc1) -- (vh);
    \draw[Arrow] (enc1.south east) -- (vt.west);
    \draw[Arrow] (enc2.east) -- (vr.west);
    \node[Model, minimum width = 20mm] at (4,1) (kbc){KBC Model};
    \draw[Arrow] (vh.east) -- (kbc.north west);
    \draw[Arrow] (vr.east) -- (kbc.west);
    \draw[Arrow] (vt.east) -- (kbc.south west);
    \node[Vect] at (5,1) (score){Score};
    \draw[Arrow] (kbc) -- (score);
    
    \end{tikzpicture}
    \caption{The diagram of our approach. Green and blue blocks represent data, grey blocks represent models, and arrows represent the data flow. Entity and relation encoders are used to map a triple $(h,r,t)$ from their names (textual representations) to their vector embeddings $(\mathbf{v}_h, \mathbf{v}_r, \mathbf{v}_t)$. These vectors are used as the input to a KBC algorithm of choice to compute the score of the triple.}
    \label{model_diagram}
\end{figure*}
\section{Model for Transfer Learning}

In this section, we introduce the architecture of our model and how a pre-trained model is used to initialize the model for fine-tuning.

The model consists of two encoders, one for entities and one for relations, and a KBC model.
Given a triple $\langle h,r,t\rangle$, the entity encoder is used to map the head $h$ and the tail $t$ into their vector embeddings $\mathbf{v}_h$ and $\mathbf{v}_t$, while the relation encoder is used to map the relation $r$ into its vector embedding $\mathbf{v}_r$.
These are then used as the input to the KBC algorithm of choice to predict their score (correctness), using the loss function, as defined by the KBC model.
The two parts of the model are architecturally independent of each other and will be described in the following paragraphs.
An illustration of our approach is given in Figure~\ref{model_diagram}.

\subsection{Encoders}
\label{ModelEncodersSection}

We use two types of mappings from an entity to a low-dimensional vector space.
The first approach is to assign each entity and relation its own embedding, initialized randomly and trained jointly with the model.
This is the default approach used by most KBC models, however, to distinguish it from the RNN-based approach, we denote it \emph{NoEncoder}.

The second approach that we test is to use an RNN-based mapping from the textual representation of an entity or relation (name) to its embedding.
We use the GloVe word embeddings~\cite{glove} to map each word into a vector, and then use them as the input to the entity encoder, implemented as a GRU~\cite{gru}.
To separate it from the NoEncoder, we call this the GRU encoder throughout this paper.
\citet{olpbench} test alternative encoders as well, but find that RNN-based approaches (LSTMs, in their case) perform the most consistently across the experiments.

The size of the NoEncoder grows linearly with the number of entities, while the size of the GRU encoder grows linearly with the vocabulary.
For large knowledge bases, the latter can significantly decrease the memory usage, as the size of the vocabulary is often smaller than the number of entities and relations.

\paragraph*{Transfer between datasets.}
Here, pre-training is always done using GRU encoders, as the transfer from the NoEncoder to any other encoder requires entity matching, which we want to avoid.

When fine-tuning is done with a GRU encoder, its parameters are initialized from the pre-trained GRU parameters.
The same applies for the vocabulary, however, if the target vocabulary includes any unknown words, their word embeddings are initialized randomly.

For the initialization of the NoEncoder setup, the pre-trained GRU is used to generate initial values of all vector embeddings by encoding their textual representations.
Any unknown words are omitted, and entities with no known words are initialized randomly.

An equivalent process is used for relations.
During our preliminary experiments, we have also tried pre-training the encoders on the next-word-prediction task on English Wikipedia, however, that turned out to have a detrimental effect on the overall performance compared to randomly-initialized GRUs ($2-3\%$ MRR drop and slower convergence).
That line of experiments was not continued.

\subsection{Knowledge Base Completion Models}

We use three models for knowledge base completion, ConvE~\cite{conve}, Tuck\-ER~\cite{tucker}, and 5$^\star$E~\cite{5stare}, chosen for their strong performance on various KBC benchmarks.
In the following paragraphs, we briefly introduce the models.
We assume that vectors $\mathbf{v}_h$, $\mathbf{v}_r$, and $\mathbf{v}_t$, obtained with encoders of any type, are the input to these models.

TuckER  assigns a score to each triple by multiplying the vectors with a core tensor $\mathcal{W} \in \mathbb{R}^{d_e\times d_e \times d_r}$, where $d_e$ is the dimension of entities, and $d_r$ is the dimension of relations.
Throughout our work, we make the simplifying assumption that $d_e=d_r$ to reduce the number of hyperparameters.
During transfer, $\mathcal{W}$ from the pre-trained model is used to initialize $\mathcal{W}$ in the fine-tuned model.

ConvE assigns a score to each triple by concatenating $\mathbf{v}_h$ and $\mathbf{v}_r$, reshaping them into a 2D matrix, and passing them through a convolutional neural network (CNN).
The output of this CNN is a $d_e$-dimensional vector, which is multiplied with $\mathbf{v}_t$ and summed with a tail-specific bias term $b_t$ to obtain the score of the triple.
During transfer, the parameters of the pre-trained CNN are used as the initialization of the CNN in the fine-tuned model.
Bias terms of the fine-tuned model are initialized at random, since they are entity-specific.

5$^\star$E models consider $\mathbf{v}_h$ and $\mathbf{v}_t$ to be complex projective lines and $\mathbf{v}_r$ a vector of $2\times 2$ complex matrices.
These correspond to a relation-specific M\"{o}bius transformation of projective lines.
We refer the reader to the work of~\citet{5stare} for the details.
Unlike in ConvE and TuckER, there are no shared parameters between different relations and entities.
Pre-training thus only serves as the initialization of the embeddings.

During the time of evaluation, the model is given a triple with a missing head or tail and is used to rank all the possible entities based on how likely  they  appear in place of the missing entity.
Following~\citet{conve}, we transform head-pre\-dic\-tion samples into tail-prediction samples by introducing reciprocal relations $r^{-1}$ for each relation and transforming $\langle ?,r,t\rangle$ into $\langle t,r^{-1},?\rangle$.
Following~\citet{care}, the name of the reciprocal relation is created by adding the prefix ``inverse of''.

During our preliminary experiments, we have also experimented with BoxE~\cite{boxe}, however, we have decided not to use it for further experiments, since it was much slower to train and evaluate, compared to other models.
A single round of training of BoxE with GRU encoders on OlpBench takes over $24$ days, which we could not afford.
\section{Datasets and Experimental Setup}

This section describes the used data, the experimental setup, and the baselines.

\subsection{Datasets}
\label{Datasets}
\begin{table*}[ht!]
\centering
\begin{tabular}{lccccc}
 & OlpBench & ReVerb45K & ReVerb20K & FB15K237 & WN18RR \\ \hline
 \#entities & $2.4M$ &$27K$&$11.1K$&$14.5K$ & $41.1K$ \\ 
 \#relations & $961K$ & $21.6K$ & $11.1K$ & $237$ & $11$\\ 
 \#entity clusters & N/A & $18.6K$ & $10.8K$ & N/A & N/A\\ \hline 
 \#train triples & $30M$ & $36K$ & $15.5K$ & $272K$ & $86.8K$ \\
 \#valid triples & $10K$ &$3.6K$&$1.6K$&$17.5K$ & $3K$ \\ 
 \#test triples & $10K$ & $5.4K$ & $2.4K$&$20.5K$ & $3K$ \\ \hline
\end{tabular}
\caption{Statistics of datasets. Only ReVerb45K and ReVerb20K  come with gold entity clusters. FB15K237 and WN18RR are canonicalized, and OlpBench is too large to allow for a manual annotation of gold clusters.}
\label{DataStatistics}
\end{table*}

We use the following five datasets to test our methods
(their statistics are given in Table~\ref{DataStatistics}):

\smallskip 
\noindent \textbf{OlpBench}~\cite{olpbench} is a large-scale OKBC dataset collected from English Wikipedia that we use for pre-training.
The dataset comes with multiple training and validation sets, however, we only use the training set with \textsc{Thorough} leakage removal and \textsc{Valid-Linked} validation set, since their data have the highest quality.

\smallskip 
\noindent\textbf{ReVerb45K}~\cite{cesi} and \textbf{Re\-Verb\-20K} (\citealp{care}, adapted from~\citealp{CanonicalizingOKB}), are small-scale OKBC datasets, obtained via the ReVerb OIE tool~\cite{reverb}.
They additionally come with gold clusters of entities, and information on which knowledge-base entities refer to the same real-world entity; this information can be used to improve the accuracy of the evaluation.
We find that $99.9\%$ of test samples in ReVerb20K and $99.3\%$ of test samples in ReVerb45K contain at least one entity or relation that did not appear in OlpBench, hence requiring out-of-distribution generalization.
Two entities or relations were considered different if their textual representations differed after the pre-processing (lowercase and removal of redundant whitespace).
If two inputs still differ after these steps, the potential canonicalization has to be performed by the model.

\smallskip 
\noindent \textbf{FB15K237}~\cite{fb15k237} and \textbf{WN18\-RR}~\cite{conve} are the most commonly used datasets for knowledge base completion, collected from Freebase and WordNet knowledge bases, respectively.
Both datasets have undergone a data cleaning to remove test leakage through inverse relations, as reported by~\citet{conve}.
We highlight that WN18RR differs from other datasets in its content.
While other datasets describe real-world entities and were often collected from Wikipedia or similar sources, WN18RR consists of information on words and linguistic relations between them, creating a major domain shift between pre-training and fine-tuning.

\subsection{Experimental Setup}

In our experiments, we observe the performance of the randomly-initialized and pre-trained \textsc{Gru} and \textsc{NoEncoder} variant of each of the three KBC models.
Pre-training was done with \textsc{Gru} encoders on OlpBench.
To be able to ensure a completely fair comparison, we make several simplifications to the models, e.g., assuming that entities and relations in TuckER have the same dimension and that all dropout rates within TuckER and ConvE are equal.
While these simplifications can result in a performance drop, they allow us to run exactly the same grid search of hyperparameters for all models, excluding human factor or randomness from the search.
We describe the details of our experimental setup with all hyperparameter choices in Appendix~\ref{ExperimentalSetupAppendix}.

\subsection{Baselines}
\label{baselines}
We compare our work to a collection of baselines, re-implementing and re-training them where appropriate.
\paragraph*{Models for Knowledge Base Completion.}
We evaluate all three of our main KBC models, ConvE, TuckER, and $5^\star$E with and without encoders. 
We include results of these models, obtained by related work and additionally compare to KBC models from related work, such as BoxE~\cite{boxe},  ComplEx~\cite{Complex}, TransH~\cite{transh}, and TransE~\cite{transe}. 
We highlight that numbers from external work were usually obtained with more experiments and broader hyperparameter search compared to experiments from our work, where we tried to guarantee exactly the same environment for a large number of models.

\paragraph*{Models for Knowledge Base Canonicalization.}
\citet{care} use external tools for knowledge base canonicalization to improve the predictions of KBC models, testing multiple methods to incorporate such data into the model.
The tested methods include graph convolution neural networks~\cite{gcn}, graph attention neural networks~\cite{gat}, and newly introduced local averaging networks (LANs).
Since LANs consistently outperform the alternative approaches in all their experiments, we use them as the only baseline of this type.

\paragraph*{Transfer Learning from Larger Knowledge Bases.}
\citet{biggraph} release pre-trained embeddings for the entire WikiData knowledge graph, which we use to initialize the TuckER model. 
We intialize the entity and relation embeddings of all entities that we can find on WikiData.
Since pre-trained WikiData embeddings are only available for dimension $d=200$, we compare them to the pre-trained and randomly initialized \textsc{NoEncoder\_TuckER} of dimension $d=200$ for a fair comparison.
We do not re-train WikiData with other dimensions due to the computational resources required. 
We do not use this baseline on KBC datasets, since we could not avoid a potential training and test set cross-contamination.
WikiData was constructed from Freebase and is linked to WordNet, the knowledge bases used to construct the FB15k237 and WN18RR datasets.

\paragraph*{Transfer Learning from Language Models.}
Pre-trained language models can be used to answer KB queries~\cite{lama,yao2019kgbert}.
We compare our results to \textsc{KG-Bert} on the FB15K237 and WN18RR datasets~\citep{yao2019kgbert} and to \textsc{Okgit} on the ReVerb45K and ReVerb20K datasets~\citep{okgit}.
Using large transformer-based language models for KBC can be slow. 
We estimate that a single round of evaluation of \textsc{KG-Bert} on the ReVerb45K test set takes over 14 days, not even accounting for training or validation.
\citet{yao2019kgbert} report in their code repository that the evaluation on FB15K237 takes over a month\footnote{\url{https://github.com/yao8839836/kg-bert/issues/8}}.
For comparison, the evaluation of any other model in this work takes up to a maximum of a couple of minutes.
Not only is it beyond our resources to perform equivalent experiments for \textsc{KG-Bert} as for other models, but we also consider this approach to be completely impractical for link prediction.

\citet{okgit} combine \textsc{Bert} with an approach by~\citet{care}, taking advantage of both knowledge base canonicalization tools and large pre-trained transformers at the same time.
Their approach is more computationally efficient, since only $\langle h,r\rangle$ are encoded with \textsc{Bert}, instead of entire triplets, requiring by a magnitude fewer passes through the transformer model.
This is the strongest published approach to OKBC.
\section{Experimental Results}
This section contains the outcome of pre-training and fine-tuning experiments.
In the second part of this section, we additionally investigate an option of zero-shot transfer.

For each model, we report its mean rank (MR), mean reciprocal rank (MRR), and Hits at 10 (H@10) metrics on the test set.
We selected $N=10$ for comparison, since it was the most consistently Hits@N metric reported in related work.
We report the Hits@N performance for other values of N, the validation set performance, the running time, and the best hyperparameters in Appendix~\ref{FullResultsAppendix}.
All evaluations are performed using the filtered setting, as suggested by~\citet{transe}.
When evaluating on uncanonicalized datasets with known gold clusters, ReVerb20K and ReVerb45K, the best-ranked tail from the correct cluster is considered to be the answer, following~\citet{care}.
The formal description of all these metrics can be found in Appendix~\ref{MetricsAppendix}.

\paragraph*{Pre-training results.}
\begin{table}
\centering
\begin{tabular}{l@{}ccc@{}}
 & MR & MRR & H@10 \\ \hline
 \textsc{Gru\_TuckER} & $\mathbf{57.2K}$ &$.053$&$.097$ \\ 
 \textsc{Gru\_ConvE} & $\mathbf{57.2K}$ & $.045$ & $.086$\\ 
 \textsc{Gru\_$5^\star$E} & 60.1K &$\mathbf{.055}$&$\mathbf{.101}$ \\ \hline
 \cite{olpbench} & -- & $.039$ & $.070$\\ \hline 
\end{tabular}
\caption{Comparison of pre-trained models on OlpBench with previously best result. The best value in each column is written in \textbf{bold}.}
\label{OlpBenchResults}
\end{table}
The performance of the pre-trained models on OlpBench is given in Table~\ref{OlpBenchResults}.
Our models obtain better scores than the previously best approach based on ComplEx~\cite{ComplexOrig}, however, we mainly attribute the improvement to the use of better KBC models.

\paragraph*{OKBC results.}
\begin{table*}
\centering
\begin{tabular}{@{}c@{}c@{\ }|ccc|ccc@{}}
& & \multicolumn{3}{c}{ReVerb20K} & \multicolumn{3}{c}{ReVerb45K} \\
 Model & Pre-trained? & MR & MRR & Hits@10 & MR & MRR & Hits@10 \\ \hline
 \multirow{2}{*}{\textsc{NoEncoder\_TuckER}} & no & $2611$ &$.196$&$.267$ & $5692$ & $.109$ & $.138$ \\
  & yes & $\mathit{303}$ &$\mathit{.379}$&$\mathit{.540}$ & $\mathit{780}$ & $\mathit{.299}$ & $\mathit{.453}$ \\ \hline
 \multirow{2}{*}{\textsc{NoEncoder\_ConvE}} & no & $1419$ &$.282$&$.380$ & $2690$ & $.232$ & $.333$ \\
  & yes & $\mathit{227}$ &$\mathit{.400}$&$\mathit{.568}$ & $\mathit{666}$ & $\mathit{.345}$ & $\mathit{.500}$ \\ \hline
 \multirow{2}{*}{\textsc{NoEncoder\_$5\star$E}} & no & $2301$ &$.228$&$.334$ & $3460$ & $.152$ & $.212$ \\
  & yes & $\mathit{780}$ &$\mathit{.249}$&$\mathit{.363}$ & $\mathit{3279}$ & $\mathit{.189}$ & $\mathit{.261}$ \\ \hline
 \multirow{2}{*}{\textsc{GRU\_TuckER}} & no & $581$ &$.364$&$.505$ & $1398$ & $.302$ & $.420$ \\
  & yes & $\mathit{245}$ &$\mathit{.397}$&$\mathit{.558}$ & $\mathit{706}$ & $\mathit{.331}$ & $\mathit{.477}$ \\ \hline
 \multirow{2}{*}{\textsc{GRU\_ConvE}} & no & $334$ &$.387$&$.540$ & $824$ & $.343$ & $.488$ \\
  & yes & $\boldsymbol{\mathit{184}}$ &$\mathit{.409}$&$\mathit{.573}$ & $\mathit{600}$ & $\mathit{.357}$ & $\mathit{.509}$ \\ \hline
 \multirow{2}{*}{\textsc{GRU\_$5\star$E}} & no & $395$ &$.390$&$.546$ & $836$ & $.357$ & $.508$ \\ 
  & yes & $\mathit{202}$ &$\boldsymbol{\mathit{.417}}$&$\boldsymbol{\mathit{.586}}$ & $\boldsymbol{\mathit{596}}$ & $\boldsymbol{\mathit{.382}}$ & $\boldsymbol{\mathit{.537}}$ \\ \hline \hline
  $\blacklozenge$ \textsc{Okgit(ConvE)} & yes$^\blacktriangle$ & $527$ & $.359$ & $.499$ & $773.9$ & $.332$ & $.464$ \\ 
  $\dagger$ \textsc{CaRe(ConvE, LAN)} & no & $973$ & $.318$ & $.439$ & $1308$ & $.324$ & $.456$ \\ 
  $\dagger$ \textsc{TransE} & no & $1426$ & $.126$ & $.299$ & $2956$ & $.193$ & $.361$ \\ 
  $\dagger$ \textsc{TransH} & no & $1464$ & $.129$ & $.303$ & $2998$ & $.194$ & $.362$ \\ \hline \hline
  \multirow{2}{*}{\textsc{NoEncoder\_TuckER}\textsubscript{$d=200$}} & no & $2855$ & $.184$ & $.248$ & $5681$ & $.109$ & $.138$ \\
  & yes &$329$ & $.369$ & $.528$ & $805$ & $.275$ & $.426$  \\ \hline
  \textsc{BigGraph\_TuckER}\textsubscript{$d=200$} & yes$^\blacktriangle$ & $1907$ & $.215$ & $.291$ & $2285$ & $.234$ & $.337$ \\ \hline 
\end{tabular}
\caption{Comparison of different models with and without pre-training on the OKBC benchmarks ReVerb20K and ReVerb45K. The scores of each model are reported with and without pre-training, with the better of the two written in \textit{italics}. Separated from the rest with two lines, are previous best results on the datasets and TuckER results with $d=200$ for a fair BigGraph comparison. The best overall value in each column is written in \textbf{bold}.
Results denoted with $\dagger$ and $\blacklozenge$ were taken from~\cite{care} and~\cite{okgit}, respectively.\\
$^\blacktriangle$ Unlike all other models with a \textit{yes} entry, \textsc{BigGraph\_TuckER} and \textsc{Okgit} were not pre-trained on OlpBench, but on pre-trained WikiData embeddings or a masked language modelling objective, respectively.}
\label{ReVerbResults}
\end{table*}

Our results of the models on ReVerb20K and ReVerb45K are given  in Table~\ref{ReVerbResults}.
All models strictly improve their performance when pre-trained on OlpBench.
This improvement is particularly noticeable for NoEncoder models, which tend to overfit and achieve poor results without pre-training.
However, when initialized with a pre-trained model, they are able to generalize much better.
$5^\star$E seems to be an exception to this, likely because there are no shared parameters between relations and entities, resulting in a weaker regularization.
GRU-based models do not seem to suffer as severely from overfitting, but their performance still visibly improves if pre-trained on OlpBench.

Finally, our best model outperforms the state-of-the-art approach by \citet{okgit} on ReVerb20K and ReVerb45K.
Even when compared to pre-trained \textsc{Gru\_ConvE}, which is based on the same KBC model, \textsc{Okgit(ConvE)} and \textsc{CaRe(ConvE,LAN)} lag behind.
This is particularly outstanding, because the \textsc{Bert} and \textsc{RoBERTa} language models, used by \textsc{Okgit(ConvE)}, received by several orders of magnitude more pre-training on unstructured text than our models, making the results more significant.

Similarly, the initialization of models with BigGraph seems to visibly help the performance, however, they are in turn outperformed by a \textsc{NoEncoder} model, initialized with pre-trained encoders instead.
This indicates that our suggested pre-training is much more efficient, despite the smaller computational cost.

\paragraph*{KBC results.}
\begin{table*}
\centering
\begin{tabular}{@{}c@{}c@{\ \ }|@{\ }c@{\ \ }c@{\ \ }c|@{\ }c@{\ \ }c@{\ \ }c@{}}
& & \multicolumn{3}{c}{FB15K237} & \multicolumn{3}{c}{WN18RR} \\
 Model & pre-trained? & MR & MRR & Hits@10 & MR & MRR & Hits@10 \\ \hline
 \multirow{2}{*}{\textsc{TuckER}} & no & $166$ &$.358$&$.545$ & $4097$ & $\mathit{.468}$ & $.528$ \\
  & yes & $\mathit{151}$ &$\mathit{.363}$&$\mathit{.550}$ & $\mathit{3456}$ & $.467$ & $\mathit{.529}$ \\ \hline
 \multirow{2}{*}{\textsc{ConvE}} & no & $212$ &$.320$&$.504$ & $6455$ & $.429$ & $.479$ \\
  & yes & $\mathit{200}$ &$\mathit{.325}$&$\mathit{.510}$ & $\mathit{5792}$ & $\mathit{.435}$ & $\mathit{.486}$ \\ \hline 
 \multirow{2}{*}{\textsc{$5\star$E}} & no & $152$ &$.353$&$.539$ & $\mathit{2450}$ & $\mathit{.492}$ & $\mathit{.583}$ \\
  & yes & $\boldsymbol{\mathit{143}}$ &$\mathit{.357}$&$\mathit{.544}$ & $2636$ & $\mathit{.492}$ & $.582$ \\ \hline \hline
  \cite{conve} \textsc{ConvE} & no & $244$ & $.325$ & $.501$ & $4187$ & $.43$ & $.52$ \\
  \cite{OldDog} \textsc{ConvE} & no & -- & $.339$ & $.536$ & -- & $.442$ & $.504$ \\
  \cite{tucker} \textsc{TuckER} & no & -- & $.358$ & $.544$ & -- & $.470$ & $.526$ \\ 
  \cite{5stare} \textsc{$5\star$E} & no & -- & $\boldsymbol{.37}$ & $\boldsymbol{.56}$ & -- & $\boldsymbol{.50}$ & $\boldsymbol{.59}$ \\ \hline
  \cite{yao2019kgbert} \textsc{KG-Bert} & yes & $153$ & -- & $.420$ & $\boldsymbol{97}$ & -- & $.524$ \\ 
  \cite{boxe} \textsc{BoxE} & no & $163$ & $.337$ & $.538$ & $3207$ & $.451$ & $.541$ \\ 
  \cite{Complex} \textsc{ComplEx} & no & -- & $\boldsymbol{.37}$ & $\boldsymbol{.56}$ & -- & $.48$ & $.57$ \\
  \cite{dura} \textsc{ComplEx-Dura} & no & -- & $\boldsymbol{.371}$ & $\boldsymbol{.560}$ & -- & $.491$ & $.571$ \\
  \cite{dura} \textsc{Rescal-Dura} & no & -- & $.368$ & $.550$ & -- & $\boldsymbol{.498}$ & $.577$ \\ \hline
\end{tabular}
\caption{Comparison of different models with and without pre-training on the KBC benchmarks FB15K237 and WN18RR. The scores of each model are reported with and without pre-training, with the better of the two written in \textit{italics}. Separated from the rest with two lines, we first list previous scores obtained with the ConvE, $5^\star$E, and TuckER models, followed by other well-performing models in the literature. The best overall value in each column is written in \textbf{bold}.}
\label{fb15kResults}
\end{table*}

To evaluate the impact of pre-training on larger canonicalized knowledge bases, we compare the performance of models on FB\-15\-K237 and WN18RR.
For brevity, we treat the choice of an encoder as a hyperparameter and report the better of the two models in Table~\ref{fb15kResults}. Detailed results are given in Appendix~\ref{FullResultsAppendix}.

Pre-trained models outperform their randomly initialized counterparts as well, however, the differences are usually smaller.
We believe that there are several reasons that can explain the small difference, primarily the difference in the size.
Best models on FB15K237 and WN18RR only made between $3$ and $12$ times more steps during pre-training than during fine-tuning.
For comparison, this ratio was between $250$ to $1000$ for ReVerb20K.
The smaller improvements on FB15K237 and WN18RR can also be explained by the domain shift, as already described in Section~\ref{Datasets}.


Table~\ref{fb15kResults} additionally includes multiple recently published implementations of ConvE, $5^\star$E, and TuckER, as well as other strong models in KBC.
We note that the comparison with all these models should be taken with a grain of salt, as other reported models were often trained with a much larger hyperparameter space, as well as additional techniques for regularization (e.g., \textsc{Dura}~\citep{dura} or label smoothing~\citep{tucker}) and sampling (e.g., self-adversarial sampling~\citep{boxe}).
Due to the large number of observed models and baselines, we could not expand the hyperparameter search without compromising the fair evaluation of all compared models.

In Appendix~\ref{FullResultsAppendix}, we also report the best hyperparameters for each model.
We highlight that pre-trained models usually obtain their best result with a larger dimension compared to their randomly-initialized counterparts. 
Pre-training thus serves as a type of regularization, allowing us to fine-tune larger models.
We believe that training on even larger pre-training datasets, we could obtain pre-trained models with more parameters and even stronger improvements across many KBC datasets.

\begin{figure}[t]
    \centering
    \includegraphics[width=0.45\textwidth]{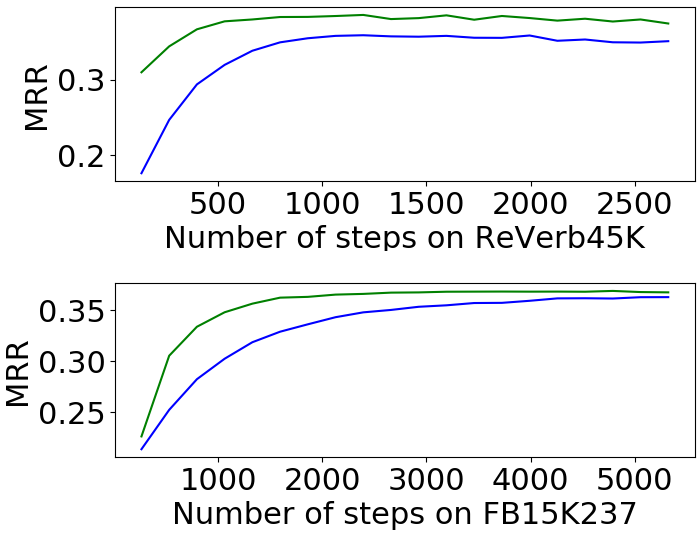}
    \caption{Comparing convergence of the overall best randomly initialized (blue) and pre-trained (green) models on ReVerb45K and FB15K237. 
    Pre-trained models converge in fewer training steps despite a smaller learning rate.}
    \label{figure_convergence}
\end{figure}

\begin{table*}[ht] 
\centering
\begin{tabular}{cc|ccc|ccc}
& & \multicolumn{3}{c}{ReVerb20K} & \multicolumn{3}{c}{ReVerb45K} \\
 Model & Pre-trained? & MR & MRR & Hits@10 & MR & MRR & Hits@10 \\ \hline
 \multirow{2}{*}{\textsc{NoEncoder\_TuckER}} & no & $4862$ &$.001$&$.002$ & $8409$ & $.001$ & $.001$ \\
  & yes & $\mathit{1486}$ &$\mathit{.012}$&$\mathit{.026}$ & $\mathit{1737}$ & $\mathit{.019}$ & $\mathit{.041}$ \\ \hline
 \multirow{2}{*}{\textsc{NoEncoder\_ConvE}} & no & $4600$ &$.002$&$\mathit{.004}$ & $8464$ & $.001$ & $\mathit{.001}$ \\
  & yes & $\mathit{1480}$ &$\mathit{.003}$&$.000$ & $\mathit{1962}$ & $\mathit{.003}$ & $\mathit{.001}$ \\ \hline
 \multirow{2}{*}{\textsc{NoEncoder\_$5\star$E}} & no & $4795$ &$.001$&$.000$ & $8445$ & $.001$ & $.002$ \\
  & yes & $\mathit{1422}$ &$\mathit{.005}$&$.000$ & $\mathit{1701}$ & $\mathit{.014}$ & $\mathit{.011}$ \\ \hline
\end{tabular}
\caption{Comparison of the zero-shot performance of different models on the OKBC benchmarks ReVerb20K and ReVerb45K. The scores of each model are reported with and without pre-training on OlpBench, with the better of the two written in \textit{italics}. Note that the model that was not pre-trained is equivalent to a random baseline.}
\label{0shotResults}
\end{table*}
Not only does pre-training allow us to train larger models, pre-trained models also require fewer training steps to obtain a similar performance, as seen in Figure~\ref{figure_convergence}.
The experiments in the figure were done with the best hyperparameter setup for both pre-trained and randomly initialized model. 
Even though the pre-trained models were fine-tuned with a smaller learning rate, they converge in fewer steps, which can be attributed to the pre-training.


Finally, we highlight that language modelling as pre-training hardly justifies the enormous computational cost.
On the majority of metrics, \textsc{KG-Bert} performs worse than models with fewer parameters and less pre-training, with a notable exception of a remarkable MR on the WN18RR dataset.

\subsection{Zero-Shot Experiments}

To understand better what kind of knowledge is transferred between the datasets, we investigate the zero-shot performance of pre-trained models on ReVerb20K and ReVerb45K, where the impact of pre-training was the strongest.
If the improvement of pre-training was mainly due to the direct memorization of facts, the zero-shot performance should already be high without fine-tuning.

The results of the zero-shot evaluation are given in Table~\ref{0shotResults}.
We report the performance of all models with and without pre-training, the latter being equivalent to the random baseline.
Note that since no fine-tuning takes place, the choice of the encoder for the  evaluation does not matter.
We thus chose to only evaluate one of them.

Observing the results, we see that pre-training the models results in a lower MR, but not necessarily a much higher MRR.
Even when the increase of MRR happens, the difference is much smaller than when comparing fine-tuned models in Table~\ref{ReVerbResults}.
This implies that the improvement induced by pre-training likely does not happen only due to the memorization of facts from OlpBench.
On the other hand, the MR of pre-trained models is comparable or even better than the MR of randomly initialized NoEncoder models, fine-tuned on the ReVerb datasets, reported in Table~\ref{ReVerbResults}.
Hence, pre-trained models carry a lot of ``approximate knowledge'', which is consistent with earlier remarks on pre-training serving as a type of regularization.

Knowing that the ReVerb20K and ReVerb45K test sets consist of facts that contain at least one previously unseen entity or relation, this can be seen as out-of-distribution generalization.
Comparing the zero-shot MRR results with the OlpBench results implies that while OKBC models are capable of out-of-domain generalization to unseen entities and relations, there is still space for improvement.
\section{Related Work}
In recent years, there have been several approaches to improve the performance of KBC algorithms through data augmentation, commonly through various levels of connection with unstructured text.
\citet{NTN}, for example, use pre-trained word embeddings to initialize their entity embeddings.
\citet{dkrl} make use of an encoder that generates an embedding given a description of the entity, and they show that their approach generalizes even to previously unseen entities.
\citet{yao2019kgbert} make use of a large-scale pre-trained transformer model to classify whether a fact is true.
They rely on costly pre-training and do not generate embeddings for entities that could be used to incorporate background knowledge into natural language understanding systems~\cite{ernie}.

Previous attempts at open knowledge base completion are tied to existing work on the canonicalization of knowledge bases.
To canonicalize open knowledge bases, automatic canonicalization tools cluster entities using manually defined features~\cite{CanonicalizingOKB} or by finding additional information from external knowledge sources~\cite{cesi}.
\citet{care} use clusters obtained with these tools to augment entity embeddings for KBC.
We note that~\citet{care} use RNN-based encoders to encode relations, but not to encode entities.
\citet{olpbench}, on the other hand, introduce a model with RNN-based encoders for both entities and relations, similarly to our approach, however, they do not transfer beyond the introduced OlpBench dataset.
Finally, \citet{okgit} use both KB canonicalization tools and large-scale pre-trained model \textsc{Bert}, combining their predictions to make a more informed decision.

The development of methods that improve predictions of KBC algorithms through data augmentation or transfer is tied to the advances in  OIE and KBC methods.
However, these are beyond the scope of this project; see the works by~\citet{oiesurvey} and~\citet{OldDog} for an overview.
\section{Summary and Outlook}
In this work, we have introduced a novel approach to transfer learning between various knowledge base completion datasets. 
The main strength of the introduced method is the ability to benefit from pre-training on uncanonicalized knowledge bases, constructed from facts, collected from unstructured text.
Scaling the introduced method up would let us train large-scale pre-trained models that have already shown to be incredibly successful in natural language processing.
We tested our method on $5$ different datasets, showing that pre-training improves the performance of models.
Pre-training turned out to be particularly beneficial on small-scale datasets, where we were able to obtain the most significant gains, e.g., $6\%$ absolute increase of MRR and $65\%$ decrease of MR over the previously best method on ReVerb20K, despite not relying on large pre-trained models like \textsc{Bert}.
There are several directions of future work, such as scaling of pre-training to larger models and datasets and investigating the impact of the encoder architecture. 

\section*{Acknowledgments}
The authors would like to thank Ralph Abboud for his helpful comments on the paper manuscript.
This work was supported by the Alan Turing Institute under the EPSRC
grant EP/N510129/1, the AXA Research Fund, the ESRC
grant “Unlocking the Potential of AI for English Law”, and
the EPSRC Studentship OUCS/EPSRC-NPIF/VK/1123106.
We also acknowledge the use of the EPSRC-funded Tier 2
facility JADE (EP/P020275/1) and GPU computing support
by Scan Computers International Ltd.

\bibliography{acl2021}
\bibliographystyle{acl_natbib}

\clearpage
\appendix
\section{Metrics}
\label{MetricsAppendix}
This section contains a formal description of the metrics that were used to score the models.
Let $\langle h,r,t \rangle$ be some test triple.
Given $\langle h,r,? \rangle$, the model is used to rank all entities in the knowledge base from the most suitable to the least suitable tail, so that the entity no.\ $1$ is the most likely tail according to the model.
All ranks are reported in the ``filtered setting'', which means that all other known correct answers, other than $t$, are removed from this list to reduce noise.
Let $r_t$ be the position or \textit{rank} of the correct tail in this list.
In ReVerb20K and ReVerb45K, there may be entities from the same cluster as $t$, equivalent to $t$.
In this case, $r_t$ is the lowest of their ranks.
The \textit{reciprocal rank} of an example is then defined as $\frac{1}{r_t}$.
The same process is repeated with the input $\langle ?,r,t\rangle$ and the head entity.

The \textit{mean rank (MR)} of the model is the average of all ranks across all test examples, both heads and tails.
The \textit{mean reciprocal rank (MRR)} is the average of all reciprocal ranks, both heads and tails.
The \textit{Hits@N} metric tells us how often the rank was smaller or equal to $N$.
The related literature most commonly uses $1$, $3$, and $10$ for values of~$N$, however, $5$, $30$, and $50$ are also occasionally reported.

\section{Detailed Experimental Setup}
\label{ExperimentalSetupAppendix}
Following~\citet{OldDog}, we use 1-N scoring for negative sampling and cross-entropy loss for all models.
The Adam optimizer~\cite{adam} was used to train the network.
We follow~\citet{conve} and~\citet{tucker} with the placement of batch norm and dropout in ConvE and TuckER, respectively, however, we simplify the setup by always setting all dropout rates to the same value to reduce the hyperparameter space.
To find the best hyperparameters, grid search is used for all experiments to exhaustively compare all options.
Despite numerous simplifications, we find that our re-implementations of the baselines (non-pre-trained NoEncoder models) perform comparable to the original reported values.
The experiments were performed on a DGX-1 cluster, using one Nvidia V100 GPU per experiment.

\paragraph*{Pre-training setup.}
Pre-training was only done with GRU encoders, as discussed in Section~\ref{ModelEncodersSection}.
Due to the large number of entities in the pre-training set, 1-N sampling is only performed with negative examples from the same batch, and batches of size $4096$ were used, following~\citet{olpbench}.
The learning rate was selected from \{$1\cdot 10^{-4}, 3\cdot 10^{-4}$\}, while the dropout rate was selected from \{$0.2, 0.3$\} for ConvE and \{$0.3, 0.4$\} for TuckER.
For 5$^\star$E, the dropout rate was not used, but N3 regularization was~\cite{Complex}, its weight selected from \{$0.1$,$0.03$\}. 
For TuckER, models with embedding dimensions $100$, $200$, and $300$ were trained.
We saved the best model of each dimension for fine-tuning.
For ConvE, models with embedding dimensions $300$ and $500$ were trained, and the best model for each dimension was saved for fine-tuning.
Following~\citet{care}, we use a single 2d convolution layer with $32$ channels and $3\times 3$ kernel size.
When the dimension of entities and relations is $300$, they were reshaped into $15\times20$ inputs, while the $20\times25$ input shapes were used for the $500$-dimensional embeddings.
For 5$^\star$E, models with embedding dimensions $200$ and $500$ were trained, and the best model for each dimension was saved for fine-tuning.
Following~\citet{olpbench}, we trained each model for $100$ epochs.
Testing on the validation set is performed each $20$ epochs, and the model with the best overall mean reciprocal rank (MRR) is selected.

\paragraph*{Fine-tuning setup.}
Fine-tuning is performed in the same way as the pre-training, however, the models were trained for $500$ epochs, and a larger hyperparameter space was considered.
More specifically, the learning rate was selected from \{$3\cdot 10^{-5}, 1\cdot 10^{-4}, 3\cdot 10^{-4}$\}. 
The dropout rate was selected from \{$0.2, 0.3$\} for ConvE and \{$0.3, 0.4$\} for TuckER. 
The weight of N3 regularization for the 5$^\star$E models was selected from \{$0.3,0.1,0.03$\}.
The batch size was selected from \{$512, 1024, 2048, 4096$\}.
The same embedding dimensions as for pre-training were considered.

\section{Full Results}
\label{FullResultsAppendix}
This appendix contains detailed information on the best performance of all models. Table~\ref{DetailedFullResults} contains the detailed information on the performance of the best models both on validation and test sets of the datasets.
Table~\ref{HyerparameterTable} includes information on the best hyperparameter setups and approximate training times.
Note that the given times are approximate and are strongly affected by the selection of the hyperparameters as well as external factors.
\begin{table*}
\centering
\tiny
\begin{tabular}{@{}c@{}c@{\ }|@{\ \ }c@{\ \ \ \ }c@{\ \ \ \ }c@{\ \ \ \ \ }c@{\ \ \ \ \ }c@{\ \ \ \ \ }c@{\ \ \ \ \ }c@{\ \ \ \ }c@{\ \ }|@{\ }c@{\ \ \ \ }c@{\ \ \ \ }c@{\ \ \ \ \ }c@{\ \ \ \ \ }c@{\ \ \ \ \ }c@{\ \ \ \ \ }c@{\ \ \ \ }c@{}}
  & & \multicolumn{8}{c}{OlpBench validation} & \multicolumn{8}{c}{OlpBench test} \\
 Model & Pre-trained? & MR & MRR & H@1 & H@3 & H@5 & H@10 & H@30 & H@50 & MR & MRR & H@1 & H@3 & H@5 & H@10 & H@30 & H@50  \\ \hline
 \textsc{GRU\_TuckER} & no & $\mathbf{55.6K}$ &$.058$&$.0.033$ & $.060$ & $.076$ & $.104$ & $.163$ & $.195$ & $\mathbf{57.2K}$ & $.053$ & $.029$ & $.054$ & $.070$ & $.097$ & $.155$ & $.189$ \\
 \textsc{GRU\_ConvE} & no & $57.6K$ &$.047$&$.025$ & $.048$ & $.063$ & $.090$ & $.147$ & $.179$ & $\mathbf{57.2K}$ & $.045$ & $.022$ & $.047$ & $.060$ & $.086$ & $.140$ & $.173$ \\
 \textsc{GRU\_$5\star$E} & no & $60.1K$ &$\mathbf{.060}$&$\mathbf{.034}$ & $\mathbf{.061}$ & $\mathbf{.078}$ & $\mathbf{.109}$ & $\mathbf{.171}$ & $\mathbf{.205}$ & $59.9K$ & $\mathbf{.055}$ & $\mathbf{.030}$ & $\mathbf{.056}$ & $\mathbf{.075}$ & $\mathbf{.101}$ & $\mathbf{.160}$ & $\mathbf{.194}$ \\ \hline \hline
   \multicolumn{18}{c}{}\\
& & \multicolumn{8}{c}{ReVerb20K validation} & \multicolumn{8}{c}{ReVerb20K test} \\
 Model & Pre-trained? & MR & MRR & H@1 & H@3 & H@5 & H@10 & H@30 & H@50 & MR & MRR & H@1 & H@3 & H@5 & H@10 & H@30 & H@50  \\ \hline
 \multirow{2}{*}{\textsc{NoEncoder\_TuckER}} & no &$2532$ & $.196$ & $.160$ & $.207$ & $.230$ & $.265$ & $.318$ & $.352$ & $2611$ &$.196$&$.157$ & $.212$& $.236$& $.267$& $.332$& $.362$ \\
  & yes & $354$ & $.367$& $.283$ & $.400$ & $.457$ & $.529$ & $.646$ & $.702$ & $303$ &$.367$ & $.295$ & $.412$ & $.466$ & $.540$ & $.659$ & $.714$ \\ \hline
 \multirow{2}{*}{\textsc{NoEncoder\_ConvE}} & no & $1376$ &$.279$& $.233$ &$.301$ &$.329$ &$.364$& $.423$& $.458$ & $1419$ & $.282$ & $.227$&$.308$ &$.338$ &$.380$&$.442$&$.475$ \\
  & yes & $292$ &$.392$& $.312$ & $.423$ & $.477$&$.546$&$.658$&$.706$ & $227$ & $.400$ & $.313$&$.440$& $.491$ & $.568$ &$.680$ &$.729$ \\ \hline
 \multirow{2}{*}{\textsc{NoEncoder\_$5\star$E}} & no & $2213$ &$.230$& $.180$ &$.243$ &$.279$ &$.325$& $.421$& $.468$ & $2301$ & $.228$ & $.174$&$.243$ &$.278$ &$.334$&$.429$&$.473$ \\
  & yes & $773$ &$.249$& $.194$ & $.263$ & $.299$&$.353$&$.466$&$.528$ & $780$ & $.249$ & $.191$&$.264$& $.302$ & $.363$ &$.471$ &$.532$ \\ \hline
 \multirow{2}{*}{\textsc{GRU\_TuckER}} & no & $646$ &$.344$&$.277$& $.366$ &$.409$ &$.470$ & $.564$ &$.609$ & $598$ & $.357$ & $.283$ & $.391$ & $.434$ &$.493$ &$.594$ &$.639$ \\
  & yes & $283$ &$.386$&$.305$ &$.416$ &$.465$ &$.543$ & $.660$ & $.707$ & $245$ & $.397$ & $.315$ & $.429$ &$.485$ & $.558$ & $.674$ & $.720$ \\ \hline
 \multirow{2}{*}{\textsc{GRU\_ConvE}} & no & $333$ &$.377$&$.305$ &$.405$ &$.447$ &$.516$ &$.635$& $.683$ & $334$ & $.387$ & $.305$ & $.421$ & $.471$ & $.540$ & $.648$ & $.690$ \\
  & yes & $\mathbf{229}$ & $.397$ & $.317$ & $.428$ &$.477$ & $.554$ & $.671$ &$.726$ & $\mathbf{184}$ & $.409$ & $.326$ & $.442$ & $.499$ & $.573$ &$.687$ & $.737$\\ \hline
 \multirow{2}{*}{\textsc{GRU\_$5\star$E}} & no & $430$ &$.379$& $.302$ &$.411$ &$.454$ &$.523$& $.639$& $.685$ & $395$ & $.390$ & $.311$&$.422$ &$.475$ &$.546$&$.650$&$.697$ \\
  & yes & $243$ &$\mathbf{.404}$& $\mathbf{.318}$ & $\mathbf{.440}$ & $\mathbf{.492}$&$\mathbf{.569}$&$\mathbf{.690}$&$\mathbf{.747}$ & $202$ & $\mathbf{.417}$ & $\mathbf{.330}$&$\mathbf{.455}$& $\mathbf{.512}$ & $\mathbf{.586}$ &$\mathbf{.701}$ &$\mathbf{.748}$ \\ \hline \hline
  \multicolumn{18}{c}{}\\
  & & \multicolumn{8}{c}{ReVerb45K validation} & \multicolumn{8}{c}{ReVerb45K test} \\
 Model & Pre-trained? & MR & MRR & H@1 & H@3 & H@5 & H@10 & H@30 & H@50 & MR & MRR & H@1 & H@3 & H@5 & H@10 & H@30 & H@50  \\ \hline
 \multirow{2}{*}{\textsc{NoEncoder\_TuckER}} & no &$5097$ & $.094$ & $.077$ & $.097$ & $.109$ & $.126$ & $.160$ & $.177$ & $5135$ &$.103$&$.088$ & $.106$ & $.116$ & $.131$ & $.164$ & $.182$ \\
  & yes & $839$ & $.305$& $.228$ & $.335$ & $.386$ & $.453$ & $.559$ & $.610$ & $780$ &$.299$ & $.220$ & $.332$ & $.386$ & $.453$ & $.557$ & $.607$ \\ \hline
 \multirow{2}{*}{\textsc{NoEncoder\_ConvE}} & no & $2741$ &$.235$& $.180$ & $.262$ & $.293$ & $.336$ & $.406$ & $.436$ & $2690$ & $.232$ & $.179$ & $.256$ & $.288$ & $.333$ & $.400$ & $.434$ \\
  & yes & $632$ &$.353$& $.274$ & $.384$ & $.438$ & $.506$ & $.611$ & $.662$ & $666$ & $.345$ & $.267$ & $.378$ & $.432$ & $.500$ & $.601$ & $.644$ \\ \hline
 \multirow{2}{*}{\textsc{NoEncoder\_$5\star$E}} & no & $3653$ &$.148$& $.116$ & $.151$ & $.174$ & $.210$ & $.289$ & $.324$ & $3460$ & $.152$ & $.123$ & $.152$ & $.174$ & $.212$ & $.289$ & $.327$ \\
  & yes & $3233$ &$.187$& $.151$ & $.195$ & $.222$ & $.258$ & $.320$ & $.349$ & $3279$ & $.189$ & $.153$ & $.198$ & $.224$ & $.261$ & $.325$ & $.357$ \\ \hline
 \multirow{2}{*}{\textsc{GRU\_TuckER}} & no & $1386$ &$.304$&$.242$ & $.330$ & $.369$ & $.423$ & $.504$ & $.545$ & $1398$ & $.302$ & $.242$ & $.323$ & $.365$ & $.420$ & $.506$ & $.546$ \\
  & yes & $761$ &$.336$&$.263$ & $.368$ & $.413$ & $.473$ & $.570$ & $.613$ & $706$ & $.331$ & $.258$ & $.359$ & $.412$ & $.477$ & $.575$ & $.618$ \\ \hline
 \multirow{2}{*}{\textsc{GRU\_ConvE}} & no & $776$ &$.351$&$.273$ & $.387$ & $.437$ & $.501$ & $.593$ & $.631$ & $824$ & $.343$ & $.268$ & $.374$ & $.424$ & $.488$ & $.584$ & $.626$ \\
  & yes & $\mathbf{584}$ & $.364$ & $.285$ & $.400$ & $.452$ & $.518$ & $.620$ & $.665$ & $600$ & $.357$ & $.277$ & $.393$ & $.444$ & $.509$ & $.607$ & $.651$\\ \hline 
 \multirow{2}{*}{\textsc{GRU\_$5\star$E}} & no & $843$ &$.362$& $.287$ & $.394$ & $.446$ & $.511$ & $.607$ & $.649$ & $836$ & $.357$ & $.278$ & $.394$ & $.443$ & $.508$ & $.605$ & $.647$ \\
  & yes & $603$ &$\mathbf{.385}$& $\mathbf{.305}$ & $\mathbf{.421}$ & $\mathbf{.474}$ & $\mathbf{.542}$ & $\mathbf{.642}$ & $\mathbf{.683}$ & $\mathbf{596}$ & $\mathbf{.382}$ & $\mathbf{.302}$ & $\mathbf{.416}$ & $\mathbf{.467}$ & $\mathbf{.537}$ & $\mathbf{.636}$ & $\mathbf{.676}$ \\ \hline \hline
    \multicolumn{18}{c}{}\\
  & & \multicolumn{8}{c}{FB15K237 validation} & \multicolumn{8}{c}{FB15K237 test} \\
 Model & Pre-trained? & MR & MRR & H@1 & H@3 & H@5 & H@10 & H@30 & H@50 & MR & MRR & H@1 & H@3 & H@5 & H@10 & H@30 & H@50  \\ \hline
 \multirow{2}{*}{\textsc{NoEncoder\_TuckER}} & no &$160$ & $.366$ & $.276$ & $.399$ & $.464$ & $.547$ & $.675$ & $.727$ & $166$ &$.358$&$.265$ & $.393$ & $.458$ & $.545$ & $.669$ & $.725$ \\
  & yes & $142$ & $\mathbf{.369}$& $\mathbf{.277}$ & $\mathbf{.403}$ & $\mathbf{.468}$ & $\mathbf{.555}$ & $\mathbf{.680}$ & $\mathbf{.732}$ & $151$ &$\mathbf{.363}$ & $\mathbf{.269}$ & $\mathbf{.398}$ & $\mathbf{.464}$ & $\mathbf{.550}$ & $\mathbf{.678}$ & $\mathbf{.733}$ \\ \hline
 \multirow{2}{*}{\textsc{NoEncoder\_ConvE}} & no & $200$ &$.326$& $.238$ & $.356$ & $.419$ & $.508$ & $.634$ & $.690$ & $212$ & $.320$ & $.230$ & $.351$ & $.417$ & $.504$ & $.634$ & $.690$ \\
  & yes & $189$ &$.332$& $.242$ & $.363$ & $.428$ & $.513$ & $.647$ & $.703$ & $200$ & $.325$ & $.233$ & $.355$ & $.421$ & $.510$ & $.645$ & $.703$ \\ \hline
 \multirow{2}{*}{\textsc{NoEncoder\_$5\star$E}} & no & $144$ &$.358$& $.267$ & $.393$ & $.456$ & $.542$ & $.670$ & $.723$ & $152$ & $.353$ & $.260$ & $.389$ & $.454$ & $.539$ & $.666$ & $.721$ \\
  & yes & $\mathbf{137}$ &$.362$& $.270$ & $.398$ & $.462$ & $.548$ & $.672$ & $.727$ & $\mathbf{143}$ & $.357$ & $.264$ & $.393$ & $.459$ & $.544$ & $.670$ & $.727$ \\ \hline
 \multirow{2}{*}{\textsc{GRU\_TuckER}} & no & $\mathbf{137}$ &$.355$&$.262$ & $.391$ & $.454$ & $.539$ & $.671$ & $.724$ & $144$ & $.350$ & $.256$ & $.386$ & $.451$ & $.538$ & $.666$ & $.723$ \\
  & yes & $\mathbf{137}$ &$.357$&$.264$ & $.393$ & $.454$ & $.542$ & $.670$ & $.726$ & $\mathbf{143}$ & $.354$ & $.260$ & $.391$ & $.453$ & $.538$ & $.668$ & $.724$ \\ \hline
 \multirow{2}{*}{\textsc{GRU\_ConvE}} & no & $161$ &$.340$&$.249$ & $.372$ & $.433$ & $.521$ & $.653$ & $.708$ & $166$ & $.334$ & $.242$ & $.368$ & $.431$ & $.519$ & $.650$ & $.705$ \\
  & yes & $151$ & $.331$ & $.241$ & $.361$ & $.423$ & $.515$ & $.652$ & $.708$ & $157$ & $.327$ & $.237$ & $.359$ & $.422$ & $.513$ & $.646$ & $.704$\\ \hline
 \multirow{2}{*}{\textsc{GRU\_$5\star$E}} & no & $145$ &$.351$& $.256$ & $.387$ & $.453$ & $.540$ & $.669$ & $.725$ & $150$ & $.345$ & $.249$ & $.380$ & $.449$ & $.536$ & $.667$ & $.725$ \\
  & yes & $143$ &$.351$& $.257$ & $.386$ & $.451$ & $.537$ & $.667$ & $.722$ & $145$ & $.348$ & $.254$ & $.384$ & $.450$ & $.536$ & $.666$ & $.720$ \\ \hline \hline
     \multicolumn{18}{c}{}\\
  & & \multicolumn{8}{c}{WN18RR validation} & \multicolumn{8}{c}{WN18RR test} \\
 Model & Pre-trained? & MR & MRR & H@1 & H@3 & H@5 & H@10 & H@30 & H@50 & MR & MRR & H@1 & H@3 & H@5 & H@10 & H@30 & H@50  \\ \hline
 \multirow{2}{*}{\textsc{NoEncoder\_TuckER}} & no & $3701$ & $.469$ & $.437$ & $.483$ & $.499$ & $.524$ & $.571$ & $.595$ & $4097$ &$.468$&$.435$ & $.483$ & $.502$ & $.528$ & $.576$ & $.595$ \\
  & yes & $3332$ & $.470$& $.439$ & $.479$ & $.500$ & $.529$ & $.583$ & $.604$ & $3456$ &$.467$ & $.434$ & $.480$ & $.500$ & $.529$ & $.583$ & $.604$ \\ \hline
 \multirow{2}{*}{\textsc{NoEncoder\_ConvE}} & no & $6337$ &$.429$& $.403$ & $.437$ & $.455$ & $.479$ & $.512$ & $.530$ & $6455$ & $.429$ & $.404$ & $.440$ & $.455$ & $.479$ & $.512$ & $.530$ \\
  & yes & $5678$ &$.433$& $.407$ & $.442$ & $.460$ & $.483$ & $.523$ & $.542$ & $5793$ & $.435$ & $.408$ & $.444$ & $.461$ & $.486$ & $.527$ & $.545$ \\ \hline
 \multirow{2}{*}{\textsc{NoEncoder\_$5\star$E}} & no & $2527$ &$\mathbf{.492}$& $\mathbf{.450}$ & $\mathbf{.505}$ & $\mathbf{.531}$ & $.574$ & $.649$ & $\mathbf{.680}$ & $\mathbf{2450}$ & $\mathbf{.492}$ & $.448$ & $\mathbf{.507}$ & $\mathbf{.534}$ & $\mathbf{.583}$ & $\mathbf{.652}$ & $.680$ \\
  & yes & $2576$ &$\mathbf{.492}$& $\mathbf{.450}$ & $.501$ & $.528$ & $\mathbf{.575}$ & $\mathbf{.651}$ & $\mathbf{.680}$ & $2636$ & $\mathbf{.492}$ & $\mathbf{.450}$ & $.504$ & $.533$ & $.582$ & $\mathbf{.652}$ & $\mathbf{.683}$ \\ \hline
 \multirow{2}{*}{\textsc{GRU\_TuckER}} & no & $3201$ &$.456$&$.428$ & $.464$ & $.484$ & $.506$ & $.553$ & $.574$ & $3262$ & $.456$ & $.429$ & $.465$ & $.483$ & $.505$ & $.551$ & $.573$ \\
  & yes & $2637$ &$.457$&$.422$ & $.469$ & $.490$ & $.523$ & $.579$ & $.609$ & $2790$ & $.455$ & $.418$ & $.469$ & $.494$ & $.524$ & $.577$ & $.600$ \\ \hline
 \multirow{2}{*}{\textsc{GRU\_ConvE}} & no & $4376$ &$.431$&$.407$ & $.439$ & $.456$ & $.477$ & $.518$ & $.538$ & $4474$ & $.428$ & $.402$ & $.438$ & $.455$ & $.474$ & $.512$ & $.533$ \\
  & yes & $5983$ & $.400$ & $.376$ & $.410$ & $.422$ & $.444$ & $.475$ & $.493$ & $6128$ & $.399$ & $.375$ & $.409$ & $.423$ & $.441$ & $.473$ & $.491$\\ \hline 
 \multirow{2}{*}{\textsc{GRU\_$5\star$E}} & no & $\mathbf{2444}$ &$.456$& $.418$ & $.463$ & $.492$ & $.533$ & $.598$ & $.629$ & $2545$ & $.452$ & $.413$ & $.462$ & $.489$ & $.527$ & $.595$ & $.629$ \\
  & yes & $2971$ &$.426$& $.388$ & $.440$ & $.464$ & $.494$ & $.548$ & $.575$ & $3068$ & $.420$ & $.379$ & $.437$ & $.459$ & $.488$ & $.547$ & $.573$ \\ \hline \hline
\end{tabular}
\caption{Full results on both validation and test set of all datasets. In addition to the metrics reported in the paper, we also report H@N for $N\in\{1,3,5,10,30,50\}$, which appeared in related work. The best value in each column is written in \textbf{bold}.}
\label{DetailedFullResults}
\end{table*}

\begin{table*}
\centering
\scriptsize
\begin{tabular}{cc|cccccc}
  & & \multicolumn{6}{c}{OlpBench} \\
 Model & Pre-trained? & dimension & learning rate & batch & dropout & N3 weight & time \\ \hline
 \textsc{GRU\_TuckER} & no & $300$ & $1\cdot 10^{-4}$ & $4096$ & $0.3$ & -- & $5$ days\\
 \textsc{GRU\_ConvE} & no & $500$ & $1\cdot 10^{-4}$ & $4096$ & $0.2$ & -- & $5$ days\\
 \textsc{GRU\_$5\star$E} & no & $500$ & $1\cdot 10^{-4}$ & $4096$ & -- & $0.03$ &  $12$ days\\ \hline \hline
   \multicolumn{8}{c}{}\\
  & & \multicolumn{6}{c}{ReVerb20K} \\
 Model & Pre-trained? & dimension & learning rate & batch & dropout & N3 weight & time \\ \hline
 \multirow{2}{*}{\textsc{NoEncoder\_TuckER}} & no & $300$ & $3\cdot 10^{-5}$ & $512$ & $0.4$ & -- & $30$ min \\
  & yes & $300$ & $3\cdot 10^{-4}$ & $512$ & $0.3$ & -- &$30$ min \\ \hline
 \multirow{2}{*}{\textsc{NoEncoder\_ConvE}} & no & $300$ & $3\cdot 10^{-4}$ & $1024$ & $0.3$ & -- & $30$ min\\
  & yes & $500$ & $1\cdot 10^{-4}$ & $512$ & $0.2$ & -- &$30$ min \\ \hline
 \multirow{2}{*}{\textsc{NoEncoder\_$5\star$E}} & no & $200$ & $1\cdot 10^{-3}$ & $512$ & -- & $0.3$ & $30$ min\\
  & yes & $500$ & $3\cdot 10^{-4}$ & $512$ & -- & $0.03$ &$30$ min \\ \hline
 \multirow{2}{*}{\textsc{Gru\_TuckER}} & no & $300$ & $3\cdot 10^{-3}$ & $1024$ & $0.4$ & -- & $30$ min \\
  & yes & $300$ & $3\cdot 10^{-5}$ & $2048$ & $0.4$ & -- & $30$ min\\ \hline
 \multirow{2}{*}{\textsc{Gru\_ConvE}} & no & $300$ & $3\cdot 10^{-5}$ & $512$ & $0.3$ & -- & $30$ min\\
  & yes & $500$ & $3\cdot 10^{-5}$ & $512$ & $0.3$ & -- & $30$ min\\ \hline 
 \multirow{2}{*}{\textsc{Gru\_$5\star$E}} & no & $200$ & $3\cdot 10^{-4}$ & $2048$ & -- & $0.1$ & $30$ min\\
  & yes & $500$ & $1\cdot 10^{-4}$ & $1024$ & -- & $0.1$ & $30$ min\\ \hline \hline
   \multicolumn{8}{c}{}\\
  & & \multicolumn{6}{c}{ReVerb45K} \\
 Model & Pre-trained? & dimension & learning rate & batch & dropout & N3 weight & time \\ \hline
 \multirow{2}{*}{\textsc{NoEncoder\_TuckER}} & no & $100$ & $1\cdot 10^{-3}$ & $512$ & $0.4$ & -- & $2.5$h\\
  & yes & $300$ & $3\cdot 10^{-4}$ & $4096$ & $0.3$ & -- & $2.5$h\\ \hline
 \multirow{2}{*}{\textsc{NoEncoder\_ConvE}} & no & $300$ & $3\cdot 10^{-4}$ & $512$ & $0.3$ & -- & $2$h\\
  & yes & $500$ & $1\cdot 10^{-4}$ & $2048$ & $0.3$ & -- & $2.5$h\\ \hline
 \multirow{2}{*}{\textsc{NoEncoder\_$5\star$E}} & no & $200$ & $1\cdot 10^{-3}$ & $2048$ & -- & $0.3$ & $3$h\\
  & yes & $200$ & $1\cdot 10^{-4}$ & $512$ & -- & $0.1$ & $3$h\\ \hline
 \multirow{2}{*}{\textsc{Gru\_TuckER}} & no & $300$ & $1\cdot 10^{-3}$ & $512$ & $0.4$ & -- & $3$h \\
  & yes & $300$ & $3\cdot 10^{-4}$ & $2048$ & $0.4$ & -- & $3$h\\ \hline
 \multirow{2}{*}{\textsc{Gru\_ConvE}} & no & $300$ & $1\cdot 10^{-4}$ & $4096$ & $0.3$ & -- &$2$h \\
  & yes & $500$ & $3\cdot 10^{-4}$ & $2048$ & $0.3$ & -- & $2.5$h\\ \hline 
 \multirow{2}{*}{\textsc{Gru\_$5\star$E}} & no & $500$ & $3\cdot 10^{-4}$ & $1024$ & -- & $0.03$ &$3$h \\
  & yes & $500$ & $3\cdot 10^{-4}$ & $2048$ & -- & $0.1$ & $3$h\\ \hline \hline
   \multicolumn{8}{c}{}\\
  & & \multicolumn{6}{c}{FB15K237} \\
 Model & Pre-trained? & dimension & learning rate & batch & dropout & N3 weight & time \\ \hline
 \multirow{2}{*}{\textsc{NoEncoder\_TuckER}} & no & $200$ & $3\cdot 10^{-4}$ & $1024$ & $0.4$ & -- & $9.5$h\\
  & yes & $300$ & $3\cdot 10^{-5}$ & $2048$ & $0.4$ & -- & $9.5$h\\ \hline
 \multirow{2}{*}{\textsc{NoEncoder\_ConvE}} & no & $300$ & $3\cdot 10^{-4}$ & $2048$ & $0.3$ & -- &$8.5$h \\
  & yes & $500$ & $3\cdot 10^{-4}$ & $512$ & $0.3$ & -- & $9$h\\ \hline
 \multirow{2}{*}{\textsc{NoEncoder\_$5\star$E}} & no & $500$ & $1\cdot 10^{-4}$ & $512$ & -- & $0.3$ &$12$h \\
  & yes & $500$ & $1\cdot 10^{-4}$ & $512$ & -- & $0.3$ & $12$h\\ \hline
 \multirow{2}{*}{\textsc{Gru\_TuckER}} & no & $100$ & $3\cdot 10^{-4}$ & $512$ & $0.4$ & -- & $13$h\\
  & yes & $200$ & $1\cdot 10^{-4}$ & $1024$ & $0.4$ & -- & $13$h\\ \hline
 \multirow{2}{*}{\textsc{Gru\_ConvE}} & no & $300$ & $3\cdot 10^{-4}$ & $512$ & $0.3$ & -- & $11$h\\
  & yes & $500$ & $3\cdot 10^{-4}$ & $512$ & $0.3$ & -- & $12$h\\ \hline 
 \multirow{2}{*}{\textsc{Gru\_$5\star$E}} & no & $500$ & $1\cdot 10^{-4}$ & $1024$ & -- & $0.1$ & $24$h\\
  & yes & $500$ & $3\cdot 10^{-4}$ & $512$ & -- & $0.1$ & $24$h\\ \hline \hline
     \multicolumn{8}{c}{}\\
  & & \multicolumn{6}{c}{WN18RR} \\
 Model & Pre-trained? & dimension & learning rate & batch & dropout & N3 weight & time \\ \hline
 \multirow{2}{*}{\textsc{NoEncoder\_TuckER}} & no & $100$ & $1\cdot 10^{-3}$ & $512$ & $0.3$ & -- & $7.5$h\\
  & yes & $100$ & $1\cdot 10^{-3}$ & $512$ & $0.3$ & -- & $7.5$h\\ \hline
 \multirow{2}{*}{\textsc{NoEncoder\_ConvE}} & no & $300$ & $1\cdot 10^{-3}$ & $512$ & $0.3$ & -- &$8$h \\
  & yes & $300$ & $1\cdot 10^{-3}$ & $512$ & $0.3$ & -- & $8$h\\ \hline
 \multirow{2}{*}{\textsc{NoEncoder\_$5\star$E}} & no & $200$ & $1\cdot 10^{-3}$ & $512$ & -- & $0.3$ &$8$h \\
  & yes & $500$ & $1\cdot 10^{-3}$ & $1024$ & -- & $0.3$ & $8$h\\ \hline
 \multirow{2}{*}{\textsc{Gru\_TuckER}} & no & $100$ & $1\cdot 10^{-3}$ & $512$ & $0.3$ & -- & $6$h\\
  & yes & $100$ & $1\cdot 10^{-3}$ & $1024$ & $0.4$ & -- & $6$h\\ \hline
 \multirow{2}{*}{\textsc{Gru\_ConvE}} & no & $300$ & $1\cdot 10^{-3}$ & $1024$ & $0.3$ & -- & $8.5$h\\
  & yes & $500$ & $1\cdot 10^{-3}$ & $2048$ & $0.3$ & -- & $8.5$h\\ \hline 
 \multirow{2}{*}{\textsc{Gru\_$5\star$E}} & no & $500$ & $3\cdot 10^{-4}$ & $512$ & -- & $0.1$ & $8.5$h\\
  & yes & $500$ & $1\cdot 10^{-3}$ & $1024$ & -- & $0.3$ & $8.5$h\\ \hline \hline
\end{tabular}
\caption{Best hyperparameter setups and training time of best models for all datasets.}
\label{HyerparameterTable}
\end{table*}

\end{document}